# ForensicNet: Lightweight Attention-Enhanced MobileNetV2 for Automated Face Identification

**Savitha N. J.**
Department of Computer Science and Engineering, University Visvesvaraya College of Engineering (UVCE), Bangalore University, CMR Institute of Technology, Bengaluru, India
savithasnadig@gmail.com (corresponding author)

**Lata B. T.**
Department of Computer Science and Engineering, University Visvesvaraya College of Engineering (UVCE), Bangalore University, Bengaluru, India
lata_bt@yahoo.co.in

 

## ABSTRACT

**In forensic environments, automated identification of perpetrators is difficult due to pose changes, changes in light, occlusion, and lack of labeled data. This paper presents ForensicNet, a lightweight deep learning framework for forensic face recognition that enhances attention. The suggested model combines the MobileNetV2 backbone with Convolutional Block Attention Modules (CBAM) to improve the learning of discriminative features while maintaining computational speed. A two-phase transfer learning strategy with adaptive layer unfreezing is used to improve domain adaptation and reduce overfitting. This study used publicly available datasets such as LFW and SCFace, with 15,000 facial images spanning 68 identity classes. The proposed model outperforms baseline architectures such as AlexNet, ResNet-50, and MobileNetV2, with an accuracy of 92.4%, a precision of 90.8%, and a recall of 89.5%. Additionally, the framework requires only 2.1 GFLOPs per inference, and hence can be used in real-time forensic surveillance applications.**

***Keywords-forensic face recognition; forensic face identification; MobileNetV2; CBAM attention; transfer learning; lightweight deep learning; surveillance analytics***

## I. INTRODUCTION

Suspect detection is one of the cornerstones of modern surveillance, forensics, and law enforcement. Although deep learning for face recognition systems has grown in popularity, it continues to perform suboptimally in natural forensic situations. The failure to recognize an individual can be due to several reasons, such as low-quality images, poor and/or uneven lighting, occlusions (face coverings, accessories, etc.), and motion blur [1]. These situations are typical for forensic work, and poor image quality in surveillance systems only compounds the problem. Although deep residual networks and Convolutional Neural Networks (CNNs), such as AlexNet, have shown impressive results in distinguishing objects on various benchmark datasets, they are often not ideal for forensic systems due to the extensive computational resources they need. The lack of sufficient surveillance systems is also a reason for the decline of such systems in facial recognition [2]. MobileNetV2 and similar models have been developed to address the constraints concerning computing limits.

The Convolutional Block Attention Module (CBAM) adds both channel-wise and spatial attention mechanisms. This lets networks focus on important parts of the face while blocking out background noise [3]. Another major challenge in forensic face recognition is the limited availability of labeled forensic datasets. Transfer learning has therefore become an effective strategy for adapting pretrained models to specialized domains. A carefully designed training strategy is therefore necessary to balance domain adaptation and model generalization [4]. To address these challenges, this paper proposes ForensicNet, a lightweight and attention-enhanced deep learning framework for automated forensic face identification. The proposed framework integrates MobileNetV2 with CBAM attention mechanisms to improve discriminative feature learning while maintaining computational efficiency.

SphereFace [5] added angular margin loss to a hypersphere embedding space to make it easier to separate classes. On the LFW benchmark, this approach achieved an accuracy of roughly 99.4% in verification, outperforming standard softmax-based CNN models. In [6], face recognition was improved for detection under occlusions by 5-8% using a multi-task CNN with added synthetic occlusion data. In [7], it was reported that deep neural networks trained with fake forshadown face data can increase the performance of forensic face recognition by 4-7% across datasets. In [8], it was identified that most computationally heavy methods maintain near 100% accuracy on Labeled Faces in the Wild (LFW) with margin-based loss functions. ArcFace [9] achieved 99.83% verification accuracy on LFW using an additive angular margin loss to enhance feature discrimination, setting a benchmark for contemporary facial recognition systems. EfficientNet implements compound model scaling to achieve optimal trade-offs of depth, width, and resolution of a model. In [10], EfficientNet-B7 achieved 84.4% top-1 accuracy on ImageNet while demonstrating good parameter efficiency. In [11], SqueezeNet, which had 50 times fewer parameters and a model size of less than 0.5 MB, was as accurate as AlexNet. This showed that highly compressed CNN architectures are possible. In [12], a lightweight CNN framework for face recognition was resistant to occlusion in real time, achieving about 89% accuracy on embedded platforms. FDREnet [13] is a face detection and recognition pipeline that utilizes Histogram of Oriented Gradients (HOG) for face detection and a Siamese deep learning architecture trained with contrastive loss to learn discriminative facial embeddings. The learned embeddings were classified using SVM and KNN clustering, achieving recognition accuracies of 98.03%, 99.57%, and 99.39% on the Face94, Face95, and Face96 datasets, respectively.

## II. METHODOLOGY

Figure 1 illustrates the overall architecture of the proposed ForensicNet, highlighting the integration of the MobileNetV2 backbone with CBAM attention modules and the final classification pipeline.

### A. Dataset and Preprocessing

Let the forensic face dataset be represented as:

$$D = \{(x_i, y_i)\}_{i=1}^{N} \quad (1)$$

where $x_i$ denotes the $i$-th input facial image and $y_i \in \{1,2,\dots,C\}$ represents the corresponding identity label among $C$ classes. The dataset consists of $N = 15{,}000$ facial images belonging to $C = 68$ distinct identities. To ensure compatibility with the MobileNetV2 backbone, each image is resized to $224 \times 224 \times 3$ before being used for model training and evaluation. To improve the generalization capability of the model under forensic image degradations, aggressive data augmentation is applied during the training phase. The augmentation techniques include random rotation within ±30° to simulate pose variability, horizontal flipping to account for mirror symmetry in facial images, and scaling and zooming to represent variations in camera distance and image capture conditions [14]. Synthetic occlusion is also introduced by masking approximately 15–30% of the facial region to emulate real-world occlusions. A weighted cross-entropy loss function is adopted, defined as:

$$L = -\frac{1}{N}\sum_{i=1}^{N} w_{y_i} \sum_{c=1}^{C} y_{i,c} \log(\hat{y}_{i,c}) \quad (2)$$

where $y_{i,c}$ denotes the one-hot ground truth indicator for class $c$, $\hat{y}_{i,c}$ represents the predicted probability for class $c$, and $w_{y_i}$ is the class weight corresponding to the $i^{th}$ sample.

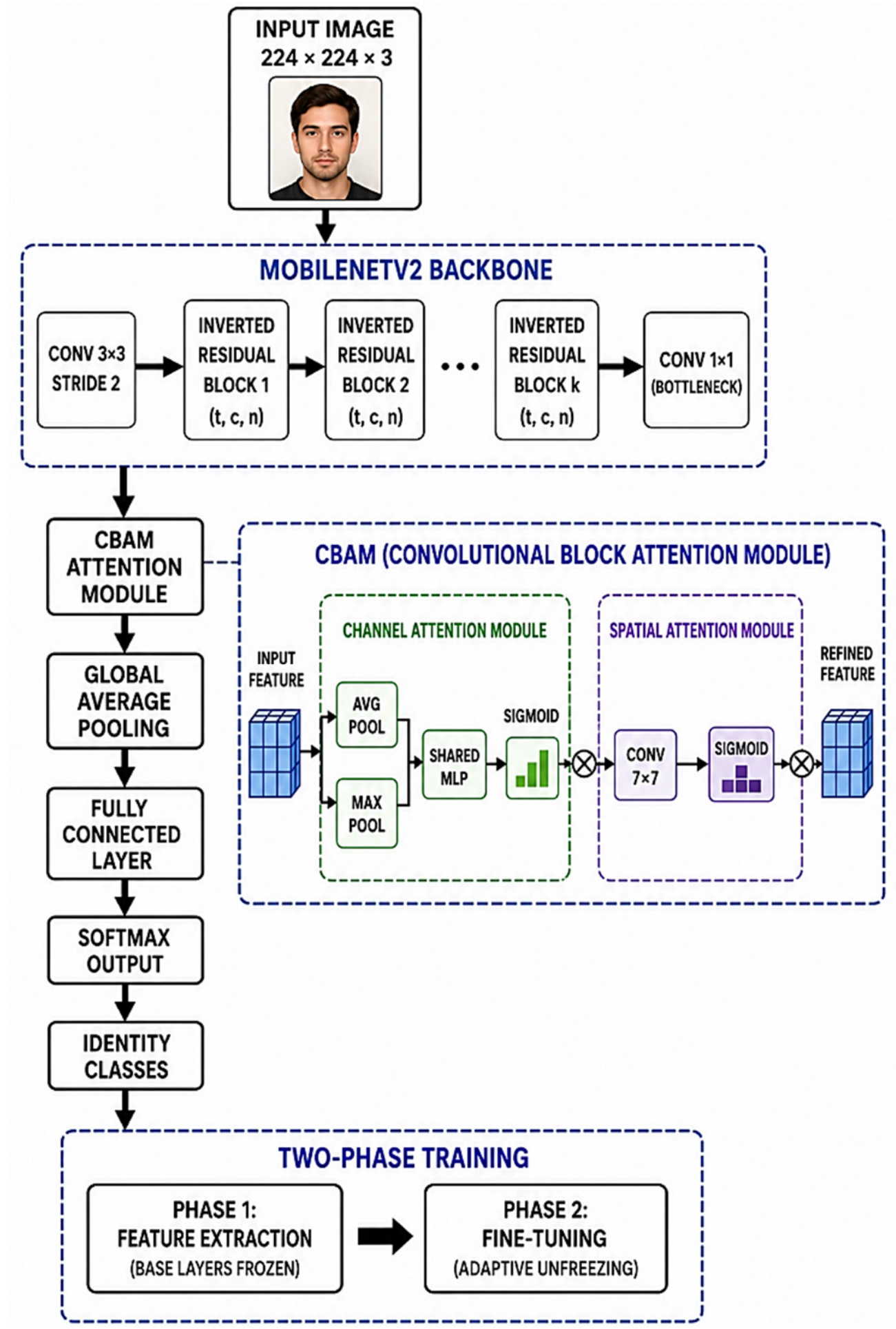

Fig. 1. Detailed architecture of the proposed ForensicNet framework integrating MobileNetV2 backbone, CBAM attention modules, adaptive transfer learning, and classification pipeline.

### B. Adaptive Layer Unfreezing Strategy

To adapt pretrained representations to the forensic domain while preventing overfitting, a two-phase transfer learning strategy with adaptive layer unfreezing is employed [15]. Let the parameters of a pretrained network be represented as

$$\theta = \{\theta_l\}_{l=1}^{L} \quad (3)$$

where $\theta_l$ denotes the parameters of the $l^{th}$ layer and $L$ represents the total number of layers. To control which layers remain trainable during transfer learning, a binary freezing mask $\phi(l)$ is defined as:

$$\phi(l) = \begin{cases} 0, & l \le l_f \\ 1, & l > l_f \end{cases} \quad (4)$$

where $l_f$ denotes the freezing threshold. When $\phi(l) = 0$, the parameters of that layer remain frozen and preserve the pretrained representations, whereas $\phi(l) = 1$ allows the layer to be updated during training. In the first phase of transfer learning, the early backbone layers are frozen ($\phi(l) = 0$) so that the network preserves general visual features such as edges, contours, and textures. During this stage, only the CBAM attention modules and the final classification layers are trained. The parameter update for trainable layers is defined as:

$$\theta_l^{t+1} = \theta_l^t - \eta \nabla_{\theta_l} \mathcal{L} \text{for } \phi(l) = 1 \quad (5)$$

where $\eta$ denotes the learning rate and $\mathcal{L}$ represents the training loss. In the second phase, deeper layers are progressively unfrozen by updating the freezing mask such that $\phi(l) = 1$ for additional layers. Fine-tuning is then performed with a reduced learning rate $\eta'$, where $\eta' < \eta$, enabling stable adaptation of high-level representations. The two-phase transfer learning strategy consists of an initial feature extraction phase, where only the classification head and attention modules are trained while backbone layers remain frozen, followed by a fine-tuning phase in which deeper layers are progressively unfrozen.

### C. MobileNetV2 Backbone with Attention

ForensicNet adopts MobileNetV2 as the backbone network due to its efficiency and suitability for edge deployment through inverted residual blocks and depthwise separable convolutions [16]. Given an input facial image

$$x \in \mathbb{R}^{224\times224\times3} \quad (6)$$

the backbone extracts a compact deep feature representation:

$$F = f(x) \in \mathbb{R}^{7\times7\times1280} \quad (7)$$

where $f(\cdot)$ denotes the MobileNetV2 feature extraction function. However, lightweight networks may distribute attention uniformly across the spatial domain, which can reduce robustness under occlusion or background clutter. The channel attention mechanism is defined as

$$A_c = \sigma(W_2\delta(W_1 \text{GAP}(F)) + W_2\delta(W_1 \text{GMP}(F))) \quad (8)$$

where $W_1$ and $W_2$ are learnable weights, $\delta(\cdot)$denotes the ReLU activation, and $\sigma(\cdot)$ represents the sigmoid function. The attention-refined feature map is then obtained as:

$$F_{att} = A_c \odot F \quad (9)$$

where $\odot$ denotes element-wise multiplication. This attention mechanism enhances discriminative facial regions while suppressing irrelevant background information, thereby improving forensic face identification performance.

### D. Hybrid Attention Injection and Optimization

ForensicNet employs a hybrid optimization strategy to improve both classification accuracy and the reliability of forensic decisions [17]. During training, a learning rate scheduling strategy is adopted to stabilize optimization. The learning rate decays exponentially during the initial training stage according to

$$\eta(t) = 0.001 \times 0.96^{\lfloor t/1000 \rfloor} \quad (10)$$

where $t$ denotes the training iteration. In the fine-tuning stage, a smaller learning rate $\eta_{fine} = \eta_{init}/10$ is used to enable stable parameter updates. To further prevent unstable updates, gradient clipping is applied as

$$\| \nabla \|_2 = \min(1.0, \| \nabla \|_2) \quad (11)$$

which constrains the gradient magnitude and avoids exploding gradients during training. This hybrid optimization framework improves the robustness and reliability of the proposed forensic identification model.

### E. Classification Head and Model Interpretability

The attention-refined feature representation $F'$ obtained from the CBAM module is passed to a lightweight classification head to perform identity prediction [18]. First, Global Average Pooling (GAP) is applied to compress the spatial feature maps into a compact feature vector, which is then processed through fully connected layers followed by a Softmax classifier. The classification process is defined as

$$y = \text{Softmax}\big(W_2 \cdot \text{ReLU}(W_1 \cdot \text{GAP}(F'))\big) \quad (12)$$

where $W_1$ and $W_2$ denote the learnable weight matrices of the fully connected layers, and $y$ represents the predicted probability distribution across $C$ identity classes. To prevent overfitting and improve generalization, Dropout with probability $p = 0.5$ and weight decay of $2 \times 10^{-4}$ is applied during training. The overall architecture consists of an input layer (224×224×3), followed by MobileNetV2 feature extraction blocks, CBAM attention modules, global average pooling, and a fully connected classification layer.

## III. RESULTS

### A. Experimental Details

The experimental dataset used in this study was constructed exclusively from two publicly available benchmark datasets: LFW (Labeled Faces in the Wild) [20, 21] and SCFace [22, 23]. Only identity classes containing at least 15 facial images were selected to ensure sufficient representation for stable training and evaluation. After preprocessing, filtering, and removal of duplicate, corrupted, and low-resolution samples (<50×50 pixels), the final curated dataset consisted of exactly 15,000 facial images spanning 68 identity classes, including 10,800 images obtained from the LFW dataset and 4,200 images collected from the SCFace surveillance dataset. Variations in pose, illumination, occlusion, and facial expressions were intentionally preserved to reflect realistic forensic surveillance conditions. The dataset was partitioned into 80% (12,000 images) for training, 10% (1,500 images) for validation, and the remaining 10% (1,500 images) for testing.

During training, a two-phase learning strategy was adopted to ensure stable optimization. In Phase 1, the learning rate was set to $\eta_1 = 10^{-3}$, while in Phase 2, the learning rate was reduced to $\eta_2 = 10^{-4}$ to enable fine-grained optimization during the progressive unfreezing of deeper layers. A fixed batch size of 32 was maintained throughout the training process. To reduce overfitting and enhance model generalization, dropout with probability $p = 0.5$ and weight decay of $2 \times 10^{-4}$ were applied during training.

### B. Results

The proposed ForensicNet framework was compared with baseline structures, AlexNet, ResNet-50, and Vanilla MobileNetV2, in terms of performance and results, as shown in

Table I. With the implementation of the two-phase transfer learning strategy, accuracy increases to 92.4%, which is 1.1% better, as shown in Figure 2. The proposed framework achieves 2.1 GFLOPs per inference, which is far better than the 3.8 GFLOPs of ResNet-50. The proposed model's accuracy is 3.7% better than ResNet-50 and 3.3% better than Vanilla MobileNetV2, as seen in Table II. Figure 3 shows the difference in recall amongst models. ForensicNet shows a higher recall, which indicates a better performance on true positive matches. Figure 4 shows the training loss curve for the proposed model.

TABLE I. PERFORMANCE COMPARISON

| Model | Accuracy (%) | Precision (%) | Recall (%) | F1-score (%) |
|---|---|---|---|---|
| AlexNet | 82.3 | 80.1 | 78.5 | 79.3 |
| ResNet-50 | 88.7 | 86.5 | 85.2 | 85.8 |
| Vanilla MobileNetV2 | 89.1 | 87.3 | 86.1 | 86.7 |
| Proposed | **92.4** | **90.8** | **89.5** | **90.1** |

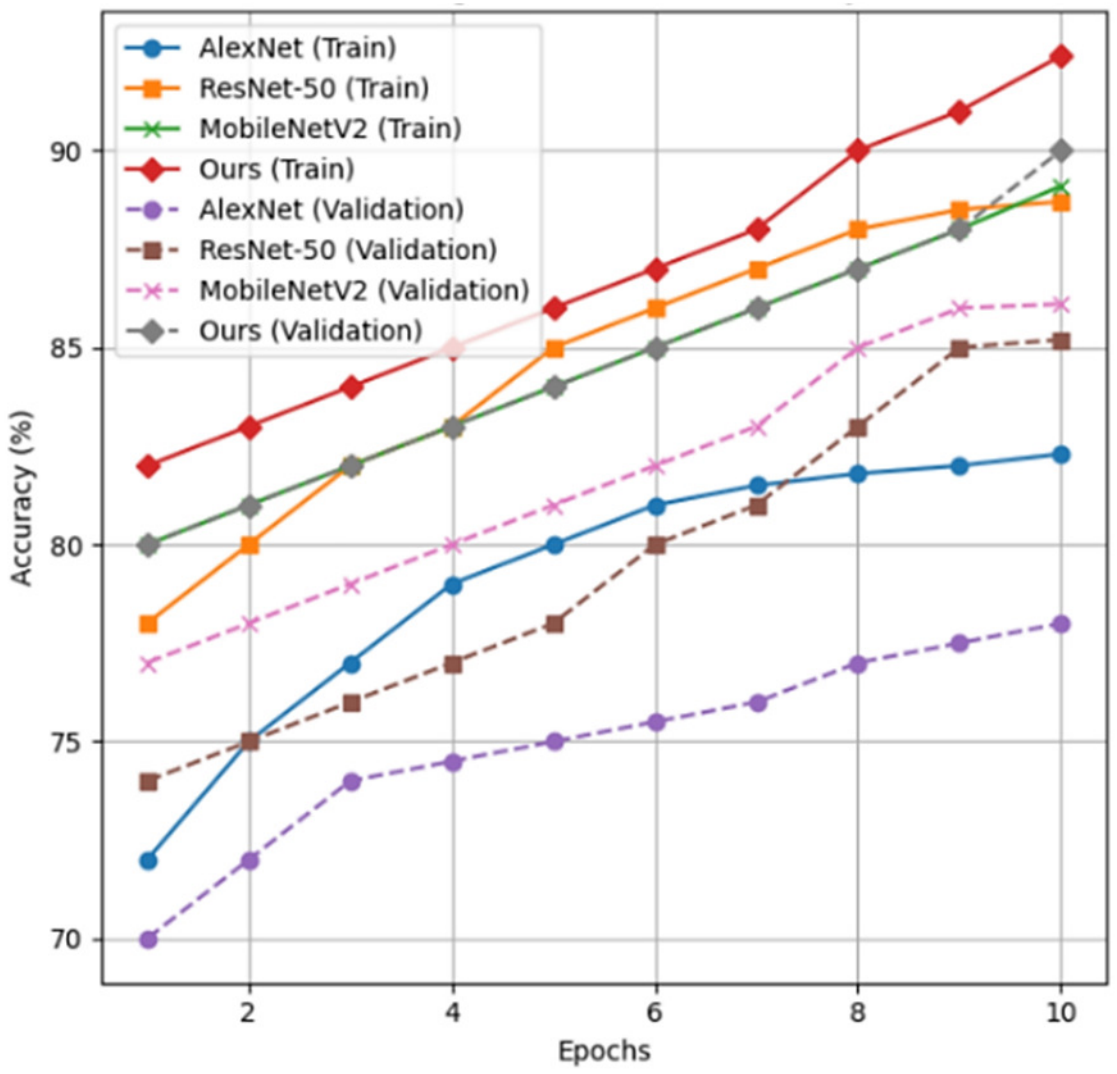

Fig. 2. Accuracy comparison.

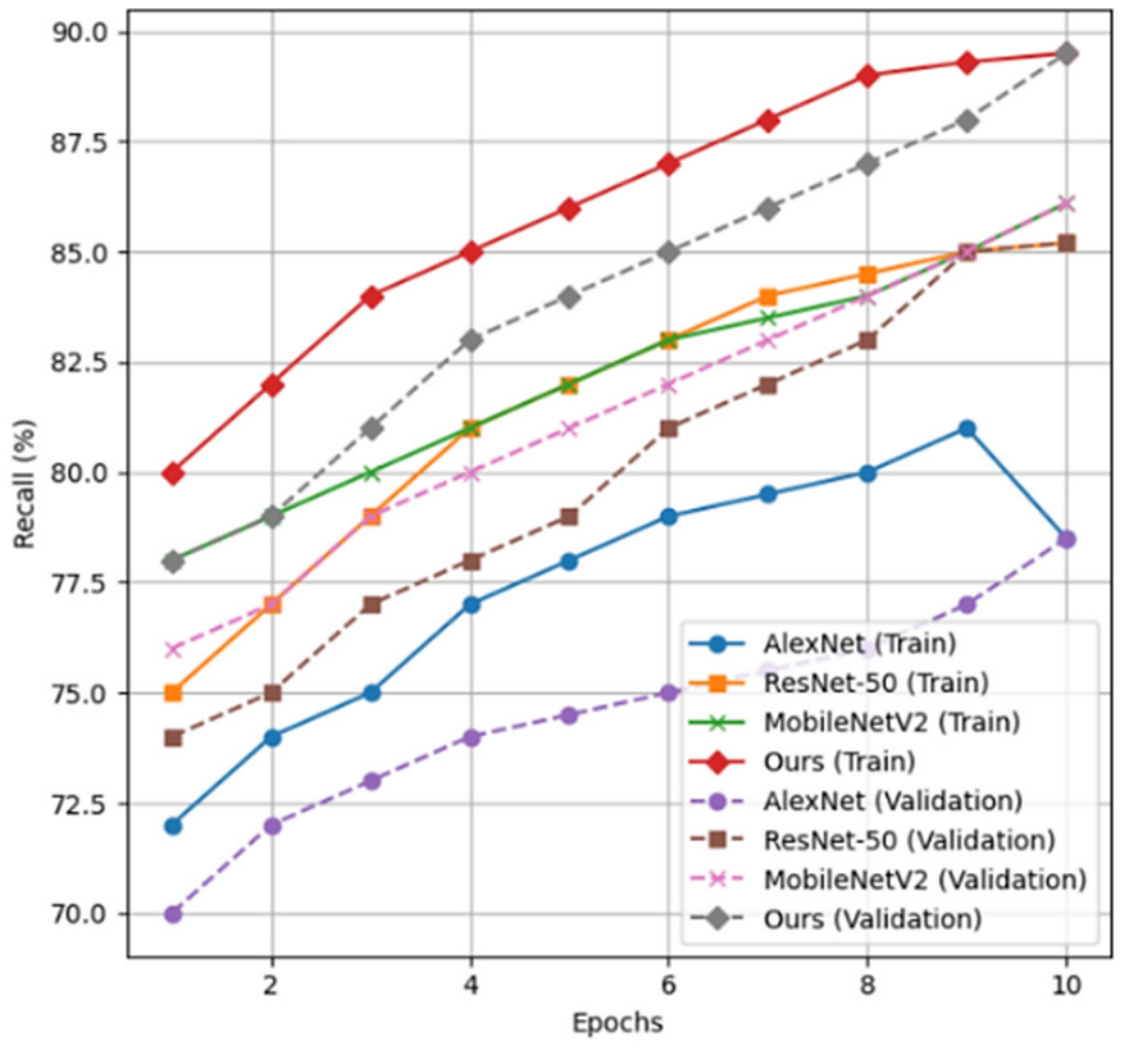

Fig. 3. Recall comparison.

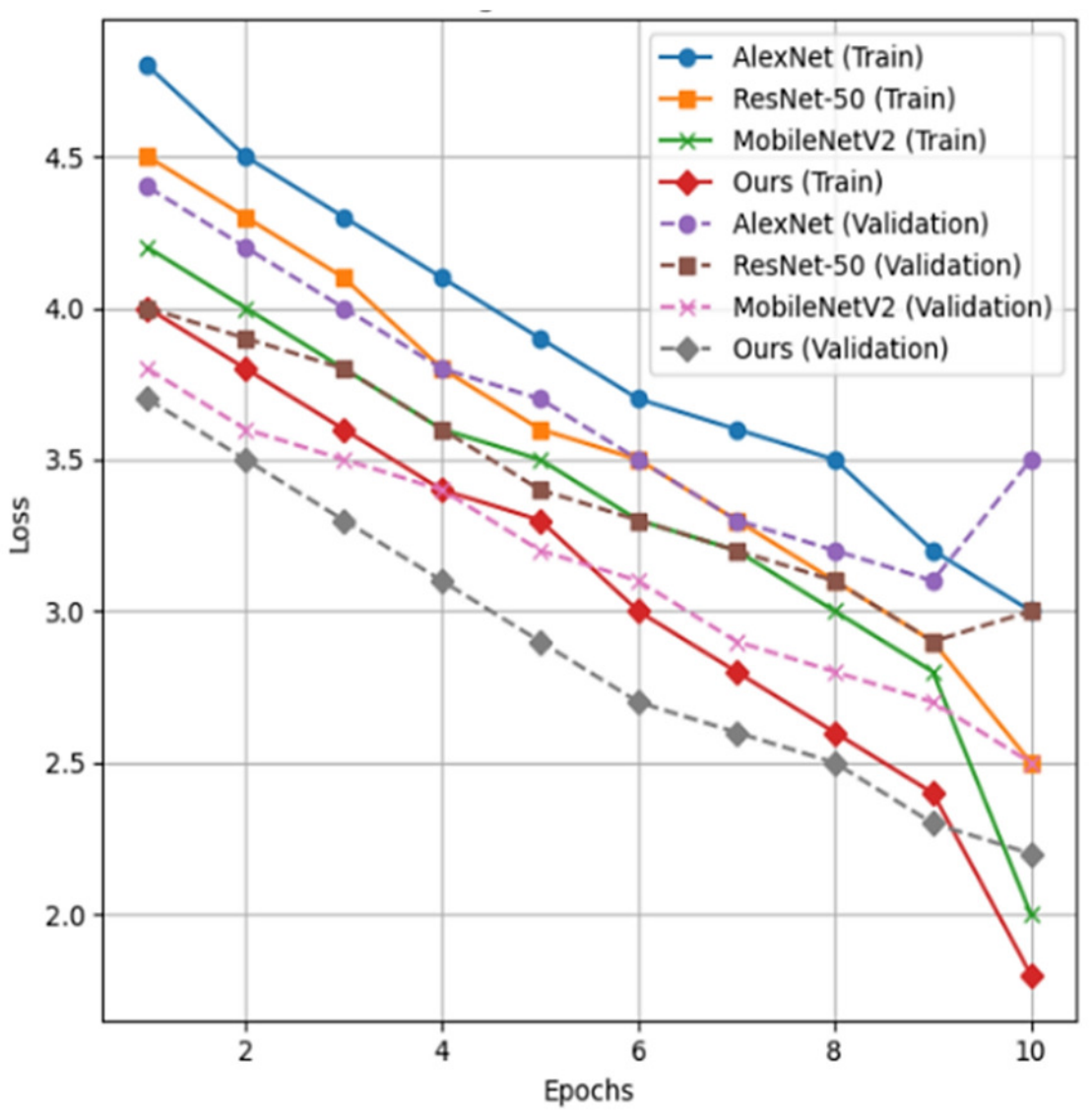

Fig. 4. Training loss comparison.

TABLE II. ABLATION STUDY

| Model Variant | Accuracy (%) |
|---|---|
| MobileNetV2 (Baseline) | 89.1 |
| + CBAM | 91.3 |
| +Two-Phase Training | **92.4** |

### *C. Comparison with Recent Deep Learning Models*

To further evaluate the effectiveness of the proposed framework, additional comparisons were conducted with recent deep learning architectures commonly used in face recognition tasks, including EfficientNet, GhostNet, and Vision Transformer (ViT)-based models. These models represent state-of-the-art approaches in lightweight CNN architectures and transformer-based visual recognition systems. To ensure a fair comparison, EfficientNet-B0, GhostNet, Vision Transformer (ViT-Small), ResNet-50, and MobileNetV2 were independently implemented and evaluated on the same curated forensic face dataset, following identical training, validation, and testing splits. Thus, the results in Table III were obtained under the same experimental protocol and are directly comparable.

TABLE III. COMPARISON WITH RECENT MODELS

| Model | Accuracy (%) | GFLOPs |
|---|---|---|
| EfficientNet-B0 | 90.6 | 3.9 |
| GhostNet | 90.2 | 2.5 |
| Vision Transformer (ViT-Small) | 91.1 | 4.6 |
| ResNet-50 | 88.7 | 3.8 |
| MobileNetV2 | 89.1 | 2.3 |
| Proposed ForensicNet | 92.4 | 2.1 |

### *D. Grad-CAM Explainability Analysis*

To facilitate understanding of the deep learning framework, Gradient-weighted Class Activation Mapping (Grad-CAM) was applied to visualize the facial areas most influential to the model's identity predictions. Grad-CAM analyzes the network's decision process for each class to create heatmaps of the most significant areas. Figure 5 illustrates anonymized Grad-CAM visualizations where only activation heatmaps are emphasized, and facial identities are masked to preserve privacy. All visualizations are presented in anonymized form by masking or blurring facial identities to ensure compliance with ethical and privacy standards.

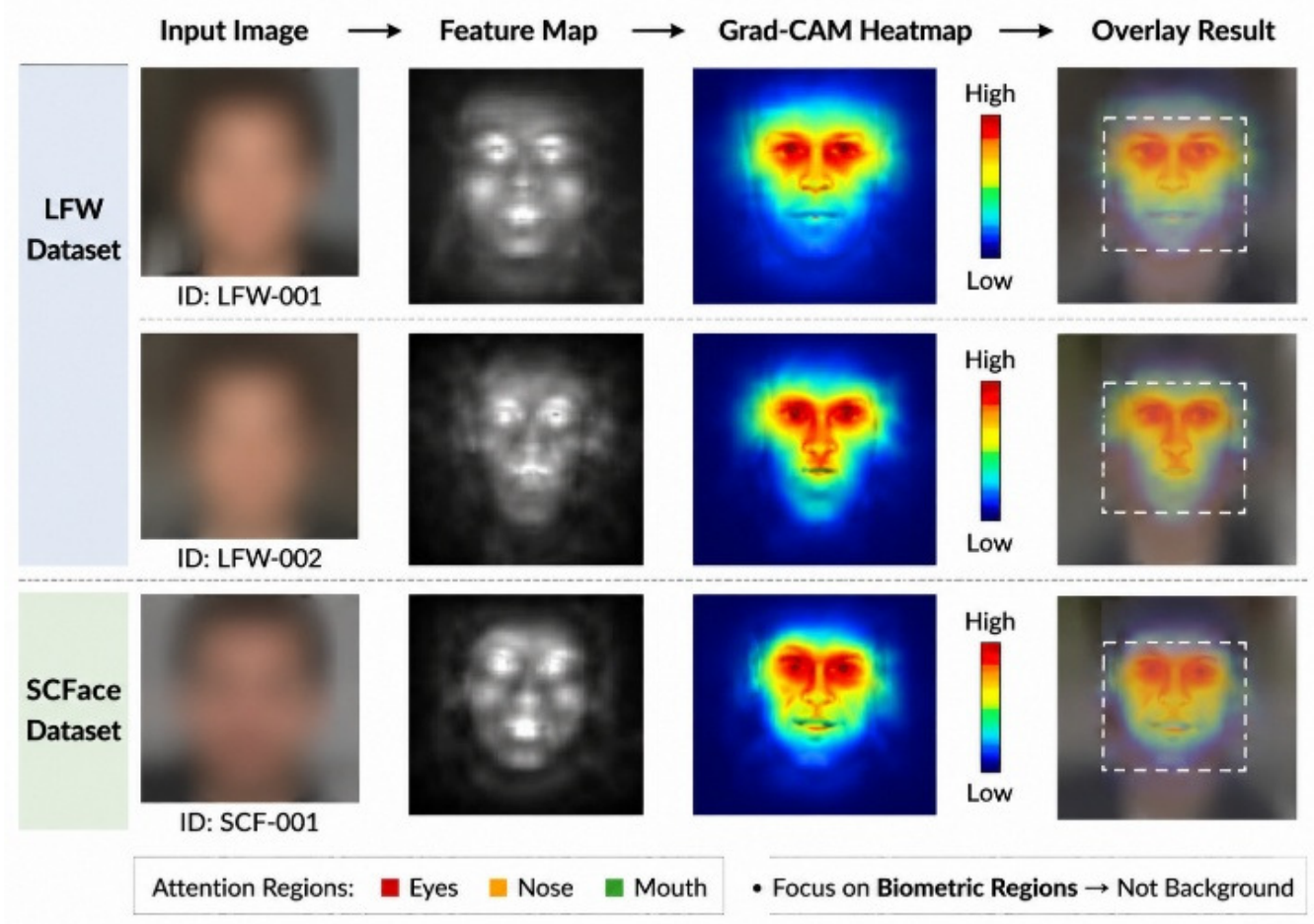

Fig. 5. Anonymized Grad-CAM visualization highlighting discriminative attention regions on forensic facial images.

## IV. CONCLUSION

This study presents ForensicNet, an efficient deep learning framework for automated forensic face identification under challenging surveillance conditions. The proposed framework integrates the MobileNetV2 backbone, CBAM, and a two-phase transfer learning strategy with adaptive layer unfreezing to address common forensic challenges such as pose variations, illumination changes, occlusion, low-resolution imagery, and limited labeled training data. The proposed framework effectively combines lightweight convolutional feature extraction with attention-enhanced discriminative learning to improve forensic face recognition under challenging surveillance conditions. Experimental evaluation conducted on a curated dataset comprising 15,000 facial images across 68 identity classes demonstrated that the proposed framework achieves 92.4% classification accuracy, outperforming conventional architectures including AlexNet, ResNet-50, and standard MobileNetV2. The ablation study further confirmed that both CBAM attention integration and adaptive transfer learning significantly contribute to improved recognition performance and model robustness.

The achieved computational complexity of only 2.1 GFLOPs demonstrates the suitability of the framework for real-time deployment in edge-based forensic surveillance systems. The lightweight architecture and reduced computational overhead make the framework suitable for deployment in practical forensic and smart surveillance applications where computational resources are limited. Despite the encouraging results, this study is limited to publicly available benchmark datasets and controlled preprocessing settings. Performance may vary under highly unconstrained real-world surveillance environments containing severe motion blur, extreme pose variations, poor illumination, and heavy occlusions. Future work will focus on integrating transformer-based lightweight architectures, evaluating larger and more diverse forensic surveillance datasets, and incorporating multispectral and low-resolution facial imagery.

## DECLARATION OF COMPETING INTERESTS

The authors declare that they have no known competing financial interests that could have appeared to influence the work reported in this paper.

## ACKNOWLEDGMENT

The authors would like to express their sincere gratitude to the Department of Computer Science and Engineering, University Visvesvaraya College of Engineering (UVCE), Bangalore University, for providing the necessary infrastructure and computational resources to carry out this research work.

## FUNDING

This research did not receive any specific grant from funding agencies in the public, commercial, or not-for-profit sectors.

## DATA AVAILABILITY

The dataset used in this study was curated from publicly available benchmark datasets, namely LFW (Labeled Faces in the Wild) [20, 21] and SCFace [22, 23]. These datasets are publicly accessible for research purposes.